\documentclass[11pt,a4paper]{article}
\usepackage[hyperref]{conll-2019}
\usepackage{amsmath, amssymb, amsthm, amscd, amsfonts}
\usepackage{times}
\usepackage{latexsym}
\usepackage{graphicx}
\usepackage{epstopdf}
\usepackage{algorithm}
\usepackage{algorithmicx}
\usepackage{algpseudocode}
\usepackage{url}
\usepackage{subfigure}
\usepackage{CJK}
\usepackage{url}
\usepackage{enumerate}
\usepackage{multirow}
\usepackage[OT1]{fontenc}

\usepackage{url}

\aclfinalcopy 

\newcommand{\toprightarrow}[1]{\mathord{\buildrel{\lower3pt\hbox{$\scriptscriptstyle\rightarrow$}}\over#1} }
\newcommand{\topleftarrow}[1]{\mathord{\buildrel{\lower3pt\hbox{$\scriptscriptstyle\leftarrow$}}\over#1} }

\title{In-Order Chart-Based Constituent Parsing}

\author{Yang Wei, Yuanbin Wu, and Man Lan \\
    School of Computer Science and Technology \\
    East China Normal University \\
    \texttt{godweiyang@gmail.com}\quad \texttt{\{ybwu,mlan\}@cs.ecnu.edu.cn}\\}

\date{}

\begin{document}
\begin{CJK*}{GBK}{song}
    \maketitle
    \begin{abstract}
        We propose a novel in-order chart-based model for constituent parsing.
        Compared with previous CKY-style and top-down models,
        our model gains advantages from in-order traversal of a tree
        (rich features, lookahead information and high efficiency) and
        makes a better use of structural knowledge by encoding the history of decisions.
        Experiments on the Penn Treebank show that our model outperforms
        previous chart-based models and achieves competitive performance
        compared with other discriminative single models.
    \end{abstract}

    \section{Introduction}
    \label{Sec:Introduction}
    Constituent parsing has achieved great progress in recent years.
    Advanced neural networks,
    such as recurrent neural networks (RNN) \citep{DBLP:conf/nips/VinyalsKKPSH15}
    and self-attentive networks \citep{DBLP:conf/acl/KleinK18},
    provide new techniques to extract powerful features for sentences.
    At the same time, new decoding algorithms are also developed to
    help searching correct parsing trees faster.
    Examples include new transition systems
    \citep{DBLP:conf/acl/WatanabeS15, DBLP:conf/naacl/DyerKBS16, DBLP:journals/tacl/LiuZ17a},
    sequence-to-sequence parsers
    \citep{DBLP:conf/emnlp/Gomez-Rodriguez18, DBLP:conf/acl/BengioSCJLS18},
    and chart-based algorithms which is our topic in this paper.

    Traditionally,
    decoders of chart parsers are based on the CKY algorithm:
    they search the optimal tree in a bottom-up manner with
    dynamic programming.
    \citet{DBLP:conf/acl/SternAK17} proposes a simple
    top-down decoder as an alternative of the CKY decoder.
    It starts at the biggest constituent (i.e., the sentence),
    and recursively splits constituents to generate
    intermediate non-terminals.
    Comparing with dynamic programming,
    the top-down parser decodes greedily
    (thus are much faster)
    and does not guarantee to reach the optimal tree.
    Furthermore,
    when predicting a non-terminal,
    the rich information about its subtrees is ignored,
    which makes the parser underperforms the CKY decoder.



    In this work, inspired by the in-order transition system \citep{DBLP:journals/tacl/LiuZ17a},
    we propose an in-order chart parser to explore local structures of a non-terminal.
    Before producing a new constituent,
    the in-order parser tries to first collect enough substructure informationy
    by predicting its left child.
    Compared with its transition-based counterpart \citep{DBLP:journals/tacl/LiuZ17a}
    which only predicts labels of a constituent
    (by shifting a non-terminal symbol into the stack),
    the in-order chart parser predicts both labels and boundaries of constituents.
    We would think that
    the additional supervision signals on boundaries can help to
    better utilize tree structures
    and also reduce search space of the parser.
    We also provide a dynamic oracle with which the in-order chart decoder
    could follow in cases of
    deviating from the gold decoding trajectory.

    Next,
    since both the in-order and top-down decoding can be seen as
    sequential decision tasks,
    we also would like to investigate whether and how previous decisions can
    influence the current decision.
    We introduce a new RNN to track history predictions,
    and the decoder will take the RNN's hidden states into account
    in each constituent prediction.

    We conduct experiments on the benchmark PTB dataset.
    The results show that the in-order chart parser can achieve competitive
    performance with the CKY and top-down decoder,
    and tracking history predictions is also useful for further improving
    the performance of decoding.




    To summarize, our main contributions include,
    \begin{itemize}
        \item A new in-order chart-based parser which is based on
              the in-order traversal of a constituent tree (Section \ref{Sec:InOrderDecoder}).
        \item A dynamic oracle for the in-order parsing
              which can provide supervision signals even in
              a wrong decoding state (Section \ref{Sec:DynamicOracle}).
        \item A mechanism based on RNNs to
              tracking history decisions in top-down and in-order parsing
              (Section \ref{Sec:StructuralRNN}).
    \end{itemize}


    \section{Encoder}
    \label{Sec:Encoder}

    \begin{algorithm}[t]
        \caption{ Top-Down Parsing. }\label{Alg:TopDownAlg}
        \begin{algorithmic}[1]
            \Function {TopDownParsing}{$i, j$}
            \If {$j = i + 1$}
            \State $\hat \ell \gets {\rm Label}(i, j)$
            \Else
            \State $\hat \ell \gets {\rm Label}(i, j)$
            \State $\hat k \gets {\rm Split}(i, j)$
            \State \Call{TopDownParsing}{$i, k$}
            \State \Call{TopDownParsing}{$k, j$}
            \EndIf
            \EndFunction
        \end{algorithmic}
    \end{algorithm}

    Given a sentence $(w_0, w_1, \ldots , w_{n-1})$ of length $n$,
    its constituent tree $T$ can be represented by
    a collection of labeled spans of the sentence, 
    \[
        \begin{aligned}
            T \triangleq \{(i, j, \ell) | & \text{ span } (i, j) \text{ with label } \ell \\
                                          & \text{ is a constituent. }\},
        \end{aligned}
    \]
    where $i$ and $j-1$ are the left and right boundary of a span respectively.
    The parsing task is to identify the spans in $T$.

    Typically, a neural constituent parser contains two components,
    an encoder which assigns scores to labeled spans
    and a decoder which finds the best span collection.
    We first describe our encoder.

    We represent each word $w_i$ using three pieces of information,
    a randomly initialized word embedding $e_i$,
    a character-based embedding $c_i$ obtained by a character-level LSTM
    and a randomly initialized part-of-speech tag embedding $p_i$.
    We concatenate these three embeddings to generate a representation of word $w_i$,
    \[
        x_i = [e_i; c_i; p_i].
    \]

    To build the representation $s_{ij}$ of an unlabeled span $(i, j)$,
    following \citep{DBLP:conf/acl/SternAK17},
    we first encode the sentence with a bidirectional LSTM.
    Let $\toprightarrow{h}_i$ and $\topleftarrow{h}_i$ be
    the forward and backward hidden states of the $i$-th position.
    The representation of span $(i, j)$ is
    the concatenation of the vector differences
    $\toprightarrow{h}_j - \toprightarrow{h}_i$ and
    $\topleftarrow{h}_i - \topleftarrow{h}_j$,
    \[
        s_{ij} = [\toprightarrow{h}_j - \toprightarrow{h}_i;
                \topleftarrow{h}_i - \topleftarrow{h}_j].
    \]

    Given $s_{ij}$, the score functions of spans and labels
    are implemented as two-layers feedforward neural networks,
    \begin{equation}
        \label{Eq:ScoringFunction}
        \begin{aligned}
            s_{\rm{label}}(i, j, \ell) & = {\bf v}_{\ell}^{\top}f({\bf W}_{\ell}^2f({\bf W}_{\ell}^1s_{ij} + {\bf b}_{\ell}^1) + {\bf b}_{\ell}^2), \\
            s_{\rm{span}}(i, j)        & = {\bf v}_s^{\top}f({\bf W}_{s}^2f({\bf W}_{s}^1s_{ij} + {\bf b}_{s}^1) + {\bf b}_{s}^2),
        \end{aligned}
    \end{equation}
    where $f$ denotes a nonlinear function (ReLU),
    and $\bf v, W, \bf b$ are model parameters.

    We define the score of a tree to be the sum of its label scores and span scores:
    \[
        s_{\rm tree}(T) = \sum\limits_{(i, j, \ell) \in T}{s_{\rm{label}}(i, j, \ell) + s_{\rm{span}}(i, j)}.
    \]

    \section{Chart-Based Decoder}
    \label{Sec:ChartBasedDecoder}

    \begin{figure*}[t]
        \centering
        \subfigure[Execution of our in-order parsing algorithm.]{
            \begin{minipage}[t]{0.58\textwidth}
                \includegraphics[width=\textwidth]{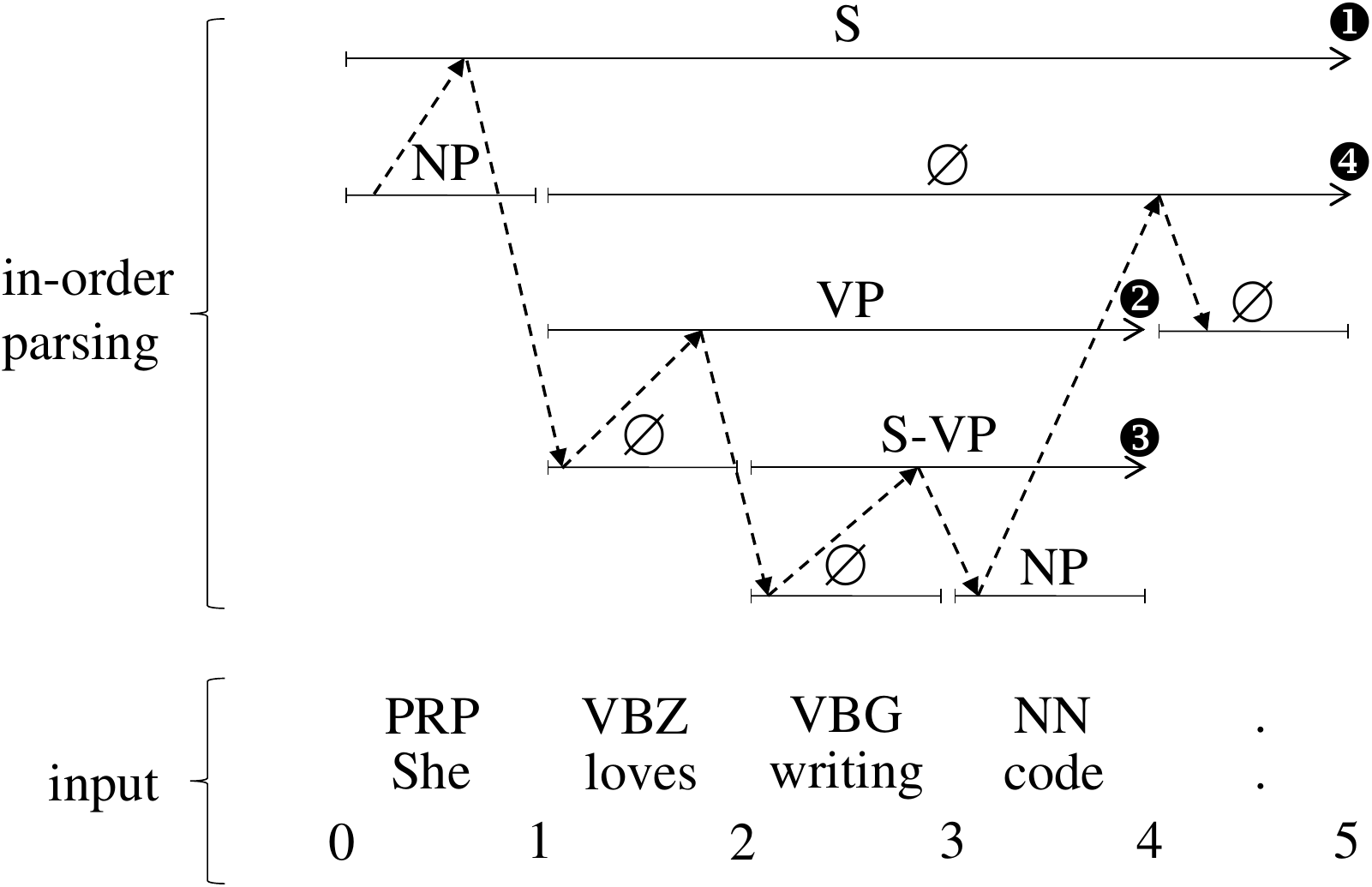}
            \end{minipage}
            \label{Fig:fig1a}
        }
        \subfigure[Output constituent tree.]{
            \begin{minipage}[t]{0.34\textwidth}
                \includegraphics[width=\textwidth]{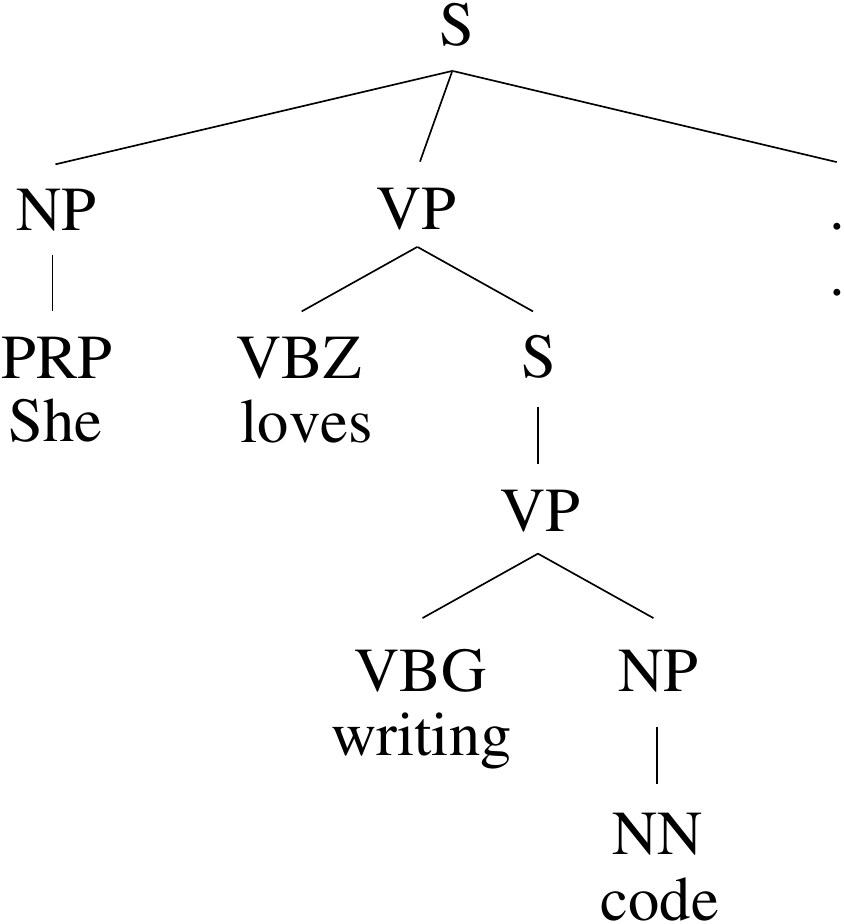}
            \end{minipage}
            \label{Fig:fig1b}
        }
        \caption{An execution of our in-order parsing algorithm (a) and the resulting constituent tree (b) for the sentence ``\emph{She loves writing code.}'' from \citet{DBLP:conf/acl/SternAK17}. Beginning with the leftmost child span $(0, 1)$, the algorithm predicts its label NP and right boundary of its parent span $5$. Because span $(0, 5)$ has reached the right bound $5$, there is no need to predict its parent span. Then the algorithm recursively acts on the right subtree ranging from $1$ to $5$ with a right bound $5$. The algorithm terminates when the parent spans of all spans have been predicted. The dotted arrows are the order in which the whole algorithm is executed and the numbers in the circles indicate the order in which the right boundaries are predicted. Notice that the empty set label $\varnothing$ represents spans which do not really exist in the gold trees and the unary label S-VP is predicted in a single step.}
        \label{Fig:fig1}
    \end{figure*}

    \begin{algorithm}[t]
        \caption{ In-Order Parsing. }\label{Alg:InOrderAlg}
        \begin{algorithmic}[1]
            \Function {InOrderParsing}{$i, j, R$}
            \If {$j = R$}
            \State $\hat \ell \gets {\rm Label}(i, j)$
            \Else
            \State $\hat \ell \gets {\rm Label}(i, j)$
            \State $\hat k \gets {\rm Parent}(i, j, R)$
            \State \Call{InOrderParsing}{$j, j + 1, k$}
            \State \Call{InOrderParsing}{$i, k, R$}
            \EndIf
            \EndFunction
        \end{algorithmic}
    \end{algorithm}

    The goal of a decoder is to (approximately)
    find a tree with the highest score.
    There are many strategies for designing a decoder.
    From the perspective of the tree traversal,
    we have two types of chart-based decoders.

    The first one is the CKY decoder which is based on dynamic programming.
    It follows the post-order traversal of a tree
    (first visiting children, then the parent),
    and is able to find optimal trees
    with a relative high time complexity $O(n^3)$.

    The second one is the top-down decoder \citep{DBLP:conf/acl/SternAK17}
    which is based on a greedy algorithm.
    It is executed according to the pre-order traversal of a tree
    (Algorithm \ref{Alg:TopDownAlg}).
    Given a span $(i, j)$, the top-down decoder first chooses
    the best non-terminal label $\ell$ for $(i, j)$,
    and predict a split $k$ to produce
    two new constituents $(i, k), (k, j)$.
    Then it recursively parses $(i, k)$ and $(k, j)$ until
    meeting spans of length one.
    The function ${\rm Label}$ in line 3 and 5 is:
    \[
        {\rm Label}(i, j) = \mathop{\arg\max}_{\ell} s_{\rm{label}}(i, j, \ell)
    \]
    and the function ${\rm Split}$ in line 6 is:
    \[
        {\rm Split}(i, j) = \mathop{\arg\max}_{i < k < j}
        s_{\rm{span}}(i, k) + s_{\rm{span}}(k, j)
        .
    \]
    The top-down decoder is faster than the CKY-style decoder (with complexity $O(n^2)$).
    On the other side, the way it predicting new constituents ignores
    substructures in $(i, k)$ and $(k, j)$,
    which may be helpful for finding the correct labels and splits.

    In this paper, we propose an in-order chart-based model to
    improve the top-down model and obtain performance comparable to
    the CKY-style model.
    Empirically, when we read a phrase, we usually notice its first word,
    then we may deduce the type of the phrase according to the observed word.
    For example, when we read the word ``\emph{loves}'',
    we may assume that it leads a verb phrase and the following words
    ``\emph{computer science}'' are just the object of the verb phrase.
    Compared with the top-down model, the local information
    in ``\emph{loves}'' may be crucial for identifying the verb phrase.
    Furthermore, the lookahead information in label VP is
    also beneficial for the next predictions.


    \subsection{In-Order Decoder}
    \label{Sec:InOrderDecoder}

    Different from the top-down parsing,
    the in-order traversal of a tree first visits its left subtree,
    then the parent and finally the right subtree.
    Algorithm \ref{Alg:InOrderAlg} shows the pseudo code
    for our in-order parsing.
    Given a span $(i, j)$,
    we first find the best label for it,
    \begin{equation}
        \label{Eq:label}
        \hat \ell = {\rm Label}(i, j).
    \end{equation}
    Then, instead of predicting a split, it predicts the parent span of $(i, j)$.
    Specifically, it seeks a position $k$ on the right of $j$,
    and $(i, k)$ is the parent of $(i, j)$,
    \begin{equation}
        \label{Eq:span}
        \hat k = {\rm Parent}(i, j, R),
    \end{equation}
    where ${\rm Parent}$ is defined as:
    \[
        {\rm Parent}(i, j, R) = \mathop{\arg\max}_{j < k \le R}
        s_{\rm{span}}(i, k) + s_{\rm{span}}(j, k).
    \]
    Note that to conduct the prediction recursively, $k$
    is constrained to be smaller than a bound $R$.
    The algorithm begins with span $(0, 1)$ and $R=n$.
    \footnote{
        For unary chains in $T$ (i.e., spans with the same boundaries but different labels),
        we treat them as a single span with a combined label.
        For $n$-ary trees, we apply implicit binarization
        which uses an empty label $\varnothing$ to
        represent the spans that do not exist in the gold tree
        but arise in the process of parsing \citep{DBLP:conf/acl/SternAK17}.
        We fix the score of label
        $\varnothing$ to $0$
        for any span $(i, j)$ in the implementation.
    }



    Figure \ref{Fig:fig1} is an example of the parsing process. For example, if the model has just predicted that the parent span of $(0, 1)$ is $(0, 5)$, then the right boundary of the parent span of $(1, 2)$ can only range from $3$ to $5$. At the beginning, the right boundary of the parent span of $(0, 1)$ ranges from $2$ to the length of the sentence $n$.

    Compared with in-order transition-based parser \citep{DBLP:journals/tacl/LiuZ17a},
    before determining the label of a non-terminal,
    the chart parser has already predicted
    boundaries of the non-terminal
    (i.e., $\mathrm{Parent}$ is executed before next $\mathrm{Label}$).
    In fact, the in-order chart parser can be seen as augmenting
    the action ``shift a non-terminal'' in the in-order transition system
    with span boundaries
    (e.g., ``shift an NP with its right bound at position $5$'').
    One rationale behind adding the boundary constraint
    is that only the in-order sequence is not sufficient
    to determine the full structure of a tree,
    we need to know the boundaries of corresponding non-terminals.
    Hence, we can consider the supervision signals in the chart-based parser
    are more accurate than those in the in-order transition-based parser.

    Finally,
    the in-order decoder is also a greedy algorithm, and
    it enjoys similar parsing speed with top-down parsers.


    \subsection{Dynamic Oracle}
    \label{Sec:DynamicOracle}

    During the training process, the decoder may output incorrect spans at some steps and the following decoding process should be able to continue based on those incorrect intermediates. In this section, we develop a dynamic oracle for the in-order parser. It helps to provide supervision signals even in a wrong decoding state.

    \begin{algorithm}[htb]
        \caption{ Dynamic Oracle for Our In-Order Parser.}\label{Alg:DynamicOracle}
        \begin{algorithmic}[1]

            \Require
            The span $(i, j)$ to be analysed and the right bound $R$ of the parent span of it;
            \Ensure
            The set $S$ of the oracle right boundaries of the parent span of $(i, j)$, in which any element $\hat j$ satisfies $j < \hat j \le R$;

            \State Identify the smallest enclosing gold constituent $(i', j')$ of span $(i, j)$ which is not equal to it;

            \If {$j' = j$}
            \State Let $j^* = R$
            \Else
            \State Let $j^* = \min(j', R)$
            \EndIf

            \State Identify the smallest enclosing gold constituent $(\tilde i, \tilde j)$ of span $(j, j^*)$;

            \If {$\tilde i + 1 = \tilde j$}
            \State Let $S = \{\tilde j\}$
            \Else
            \State Let $S = \{k \in b(\tilde i, \tilde j) | j < k \le j^*\}$
            \EndIf

            \State \Return $S$;
        \end{algorithmic}
    \end{algorithm}

    Formally, for any span $(i, j)$, the dynamic oracle aims to find an oracle label and a set of oracle right boundaries $S$. If span $(i, j)$ is in the gold tree, following the decisions in $S$, the decoder should construct the full gold tree finally. Otherwise, the best potential tree after adopting any decisions in $S$ should be the same as the best one in the current step, which means that the decisions in $S$ should not reduce any future reachable gold spans.

    For label decisions, if a span is contained in the gold tree, the oracle label is simply the label of it in the gold tree. Otherwise the oracle label is the empty label $\varnothing$.

    For right boundary decisions, given a span $(i, j)$ and a right bound $R$, our goal is to find a set of right boundaries which are not greater than $R$. Besides, we have to make sure that the optimal reachable constituent tree generated after adopting these right boundaries is consistent with that at the current step.

    Algorithm \ref{Alg:DynamicOracle} shows our dynamic oracle for right boundary decisions. Firstly in line $1$, for span $(i, j)$, the algorithm identifies the smallest enclosing gold constituent $(i', j')$ which is not equal to it. Then from line $2$ to $6$, select $j^*$ according to the value of $j'$ and $j$. Next in line $7$, for span $(j, j^*)$, identify the smallest enclosing gold constituent $(\tilde i, \tilde j)$. Finally from line $8$ to $12$, if span $(\tilde i, \tilde j)$ is of length $1$, directly return the right boundary of it. Otherwise return the set of right boundaries of its child spans which also lie inside span $(j, j^*)$, $\{k \in b(\tilde i, \tilde j) | j < k \le j^*\}$. Here $b(\tilde i, \tilde j)$ represents the set of right boundaries of the child spans of $(\tilde i, \tilde j)$. For example, if given a span $(1, 7)$ alone with its child spans $(1, 3)$, $(3, 6)$ and $(6, 7)$, we would have $b(1, 7) = \{3, 6, 7\}$.

    In our implementation, we choose the rightmost boundary in $S$ as our oracle decision. This will not affect the performance since different choices correspond to different binarizations of the original $n$-ary tree. The proof of the correctness of our dynamic oracle is similar to that in \citet{DBLP:conf/emnlp/CrossH16}. \footnote{For better understanding, we present a more detailed explanation in the supplementary material.}


    \section{Training}
    \label{Sec:Training}

    We use margin training to learn these models which has been widely used in structured prediction \citep{DBLP:conf/icml/TaskarCKG05}.

    For a span $(i, j)$ in the gold constituent tree, let $\ell^*$ represent its gold label and $k^*$ represent the gold right boundary of its parent span. Let $\hat \ell$ and $\hat k$ represent decisions made in Equation \eqref{Eq:label} and \eqref{Eq:span}. If $\hat\ell \ne \ell^*$, we define the hinge loss as:
    \[
        \max (0, 1 - s_{\rm label}(i, j, \ell^*) + s_{\rm label}(i, j, \hat \ell)).
    \]
    Otherwise we define the loss to zero. Similarly, if $\hat k \ne k^*$, we define the hinge loss as:
    \[
        \max (0, 1 - s_{\rm parent}(i, j, k^*) + s_{\rm parent}(i, j, \hat k)),
    \]
    where $s_{\rm parent}(i, j, k)$ is defined as:
    \[
        s_{\rm parent}(i, j, k) = s_{\rm span}(i, k) + s_{\rm span}(j, k)
        .
    \]

    For a single training example, we accumulate the hinge losses at all decision points. Finally we minimize the sum of training objectives on all training examples.

    Having defined the dynamic oracle for our in-order parsing model in Section \ref{Sec:DynamicOracle}, we can deal with all the spans even if they are not in the gold tree. In our implementation, we can train with exploration to increase the numbers of incorrect samples and better handle them. More specifically, we follow the decisions predicted by the model instead of gold decisions and the dynamic oracle can provide supervision at testing time.

    \begin{figure}[t]
        \centering
        \subfigure[Chain-LSTM.]{
            \begin{minipage}[t]{\linewidth}
                \includegraphics[width=\textwidth]{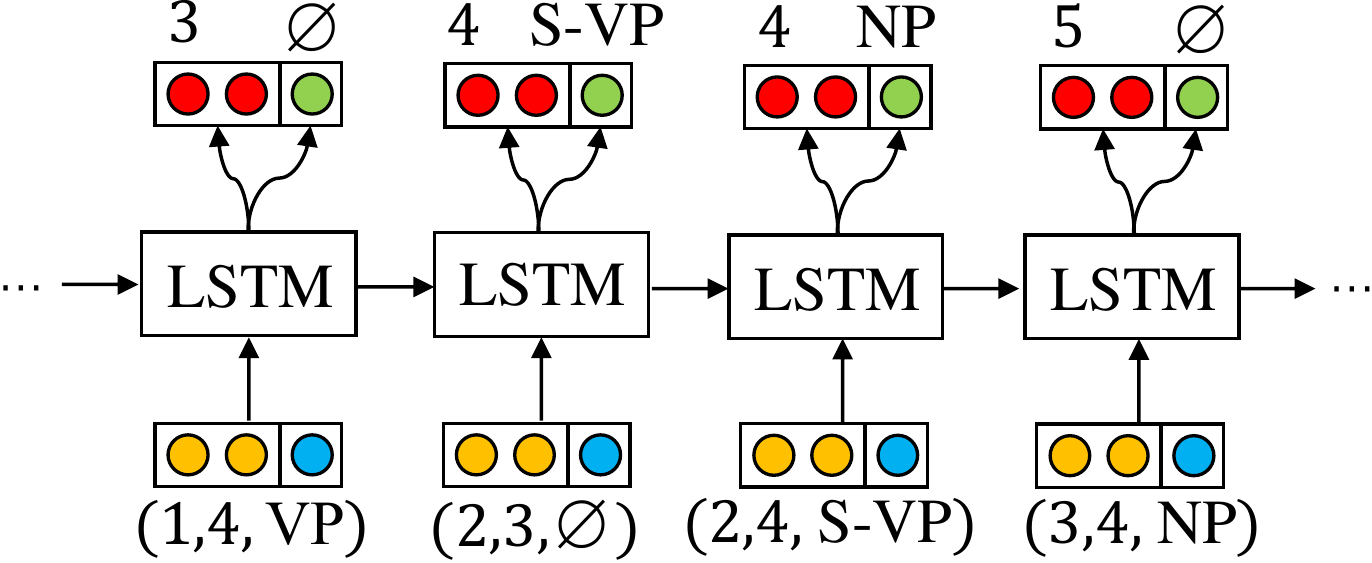}
            \end{minipage}
            \label{Fig:fig2a}
        }

        \subfigure[Stack-LSTM.]{
            \begin{minipage}[t]{\linewidth}
                \includegraphics[width=\textwidth]{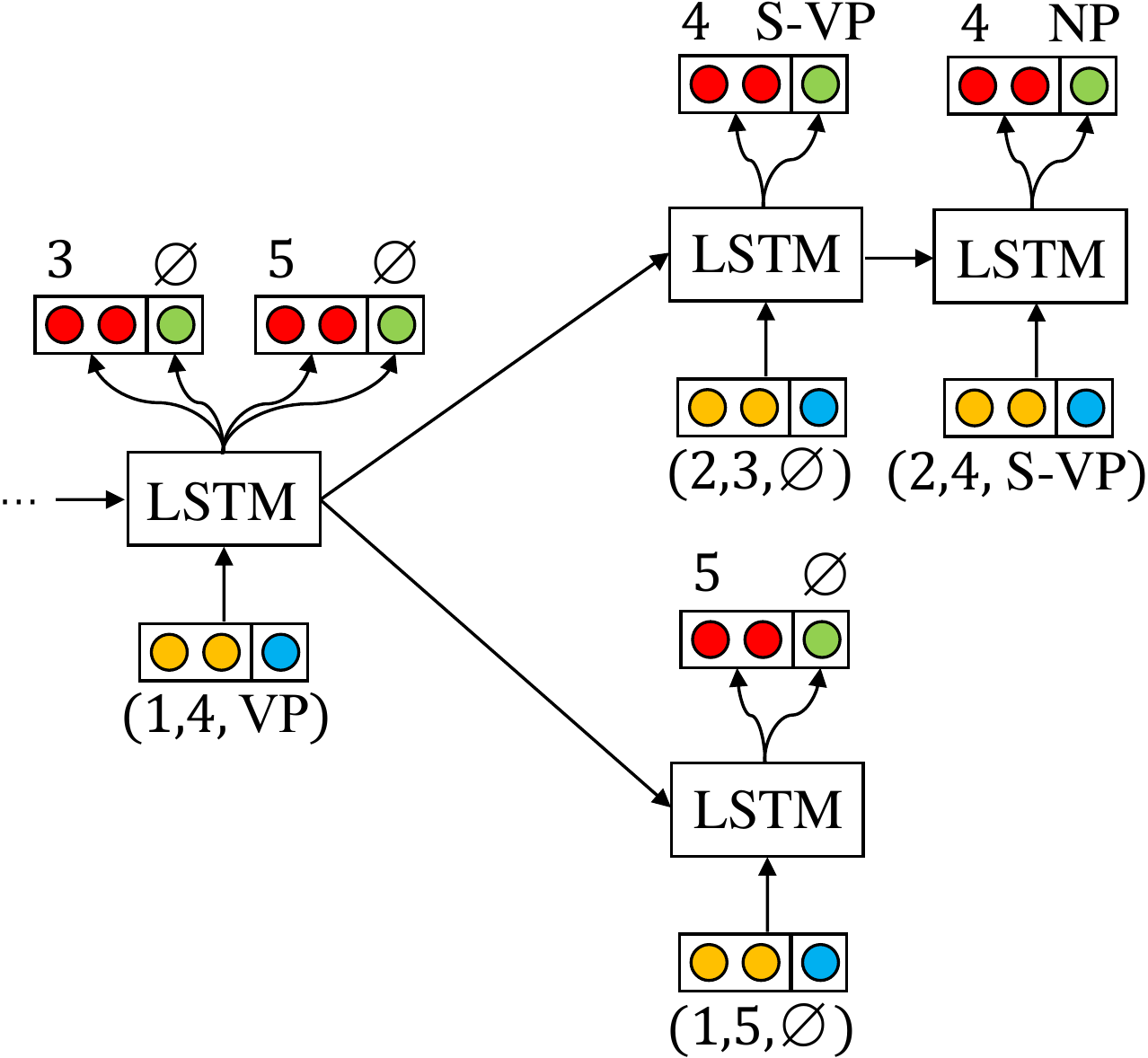}
            \end{minipage}
            \label{Fig:fig2b}
        }
        \caption{Two different types of LSTMs to encode the history of decisions in Figure \ref{Fig:fig1}. We only illustrate the decoding process of the phrase ``\emph{loves writing code.}'' here. Notice that the orange and blue input vectors denote the representations of spans and labels, respectively. The red and green output vectors denote the predictions on right boundaries and labels, respectively.}
        \label{Fig:fig2}
    \end{figure}

    \section{Tracking History Decisions}
    \label{Sec:StructuralRNN}

    Both top-down and in-order decoding
    can be seen as sequential decision processes.
    In above setting, the prediction of labels and
    parent spans (splits) only considers features of the current span.
    We can also take previous decoding decisions into account.
    Here, we propose using an LSTM to track previous decoding results
    and utilize those history information in
    the label and parent selection.

    Supposed that span $(i, j)$ with label $\ell$ is
    the $t$-th prediction of the decision sequence in our parser.
    We encode this information using a LSTM with $t$-th input
    $[s_{ij}; E_{\ell}]$ (i.e., the concatenation of
    the span representation $s_{ij}$ and the label embedding $E_{\ell}$ of $\ell$),
    \[
        h_{t} = {\rm LSTM}([s_{ij}; E_{\ell}], h_{t - 1}).
    \]
    We replace $s_{ij}$ in Equation \eqref{Eq:ScoringFunction}
    with $[s_{ij}; h_t]$ to utilize history decisions information
    in the prediction of the next labeled span.

    There are two variants of the LSTM setting.
    The first one is encoding the history decisions in the order of in-order traversal.
    As is shown in Figure \ref{Fig:fig2a},
    after predicting the labeled span $(1, 4, \text{VP})$,
    we use its representation as input of the LSTM and
    utilize the output to predict next right boundary 3,
    then its label $\varnothing$.
    We call it \emph{chain-LSTM} for that the whole LSTM is
    linear and has no branching in the middle.
    The second one is when predicting a labeled span,
    we ignore previous decisions from the right subtree of its left subtree.
    For example, in Figure \ref{Fig:fig2b},
    after inputting the representation of labeled span $(1, 4, \text{VP})$,
    we first predict next right boundary 3 with label $\varnothing$ and then predict its right subtree the same as chain-LSTM.
    However, when we predict its parent span,
    we only use the history decisions before span $(1, 4)$
    without considering its right subtree ranging from 2 to 4.
    The output of the LSTM after inputting the representation of
    labeled span $(1, 4, \text{VP})$ is used twice,
    once for predicting its right subtree, another for its parent span.
    We call it \emph{stack-LSTM} here for
    that its structure is similar to the one described in \citet{DBLP:conf/acl/DyerBLMS15}.

    \section{Experiments}
    \label{Sec:Experiments}

    \subsection{Data and Settings}
    We use standard benchmark of WSJ sections in PTB \citep{DBLP:journals/coling/MarcusSM94} for our experiments, where the Sections 2-21 are used for training data, Section 22 for development data and Section 23 for testing data.

    For words in the testing corpus but not in the training corpus,
    we replace them with a unique label \texttt{<UNK>}.
    We also replace the words in the training corpus with
    the unknown label \texttt{<UNK>} with probability
    $p_{unk}(w) = \frac{z}{z + c(w)}$, where $c(w)$ is the
    number of time word $w$ appears in the training corpus
    and we set $z = 0.8375$ as \citet{DBLP:conf/emnlp/CrossH16}.
    We use AdaDelta optimizer with epsilon $10^{-7}$ and rho 0.99.
    The other parameters are the same as the implementation of \citet{DBLP:conf/acl/SternAK17}. We use DyNet \footnote{\url{https://github.com/clab/dynet}} as our neural network toolkit and run the code on a GTX Titan GPU.

    \subsection{Results and Analysis}
    \label{Sec:Results&Analysis}

    \begin{table}[t]
        \normalsize
        \begin{center}
            \begin{tabular}{l|c|c}
                \hline
                Model                   & F1             & Sents/Sec \\
                \hline
                CKY-style parser        & 91.79          & 5.55      \\
                Top-down parser         & 91.77          & 25.70     \\
                Top-down parser$^*$     & 91.98          & 18.03     \\
                In-order S-R parser     & 91.8           & -         \\
                \hline
                Our in-order parser     & 91.87          & 23.46     \\
                Our in-order parser$^*$ & \textbf{92.03} & 17.38     \\
                \hline
            \end{tabular}
        \end{center}
        \caption{ Comparison of results and running time on the PTB test set. The in-order S-R parser represents the in-order shift-reduce transition-based parser. $^*$ represents adding the information of history decisions.}
        \label{Tab:CompTopdownEtc}
    \end{table}

    Table \ref{Tab:CompTopdownEtc} shows the comparison with two chart-based parsers in \citet{DBLP:conf/acl/SternAK17} and the in-order transition-based parser in \citet{DBLP:journals/tacl/LiuZ17a}. We implement them using the same encoder and parameter setting as our in-order parsers. Firstly, we compare our plain model with the CKY-style and top-down model. We observe that our in-order model slightly outperforms both of them in F1 score. This confirms that our model can not only consider the lookahead information of top-down model, but also benefits from the rich features of CKY-style model. Next, we compare our model with the in-order transition-based model. Our model also outperforms theirs for that their model does not predict the boundaries together with the labels, which is correspond with our conjecture that only label sequence cannot uniquely determine a constituent tree. In addition, we add the information of history decisions into the top-down parser and our in-order parser. We observe that it can improve the performance of the two parsers without losing too much efficiency. Finally, we compare the running time of these parsers. Since our in-order decoding is also based on greedy algorithm, it has similar efficiency as the top-down parser.

    In order to verify the validity of the LSTM encoding the history decisions, we remove it and parse the tree only using span representations. The comparison results are shown in Table \ref{Tab:CompPlainInOrder}. We observe that the in-order model indeed benefits from the chain-LSTM. However, adding the stack-LSTM will reduce performance unexpectedly. We speculate that stack-LSTM does not encode the whole history decisions as the chain-LSTM. It only utilizes part of the history information to guide the prediction of spans, which leads to performance degradation.

    \begin{table}[t]
        \normalsize
        \begin{center}
            \begin{tabular}{l|ccc}
                \hline
                Model                 & LR    & LP    & F1             \\
                \hline
                Plain in-order parser & 91.85 & 92.67 & 92.26          \\
                \hline\hline
                + chain-LSTM          & 91.74 & 92.95 & \textbf{92.34} \\
                + stack-LSTM          & 91.51 & 92.90 & 92.20          \\
                \hline
            \end{tabular}
        \end{center}
        \caption{ Development set results of the in-order parsers with and without the information of history decisions. LR, LP represent labeled recall and precision, respectively.}
        \label{Tab:CompPlainInOrder}
    \end{table}

    \begin{table}[t]
        \normalsize
        \begin{center}
            \begin{tabular}{l|l|ccc}
                \hline
                \multicolumn{2}{c|}{Model} & LR     & LP    & F1                     \\
                \hline
                LEmb                       & LPre   & 91.51 & 92.90 & 92.20          \\
                \hline\hline
                -                          & + SPre & 91.51 & 92.97 & 92.23          \\
                + SRep                     & -      & 91.56 & 92.97 & 92.26          \\
                + SRep                     & + SPre & 91.67 & 93.00 & \textbf{92.33} \\
                \hline
                SRep                       & LPre   & 91.27 & 93.27 & 92.26          \\
                \hline\hline
                -                          & + SPre & 91.56 & 93.02 & 92.28          \\
                \hline
            \end{tabular}
        \end{center}
        \caption{ Development set results of the in-order parsers with different configurations on the LSTM encoding history decisions. ``LEmb'' and ``SRep'' represent using label embeddings and span representations as the input of the LSTM, respectively. ``LPre'' and ``SPre'' represent using the history decisions information to predict labels and spans, respectively. All the models track the history decisions based on the stack-LSTM here.}
        \label{Tab:CompDiffConfig}
    \end{table}

    \begin{table}[t]
        \normalsize
        \begin{center}
            \begin{tabular}{l|ccc}
                \hline
                Model                                 & LR   & LP   & F1            \\
                \hline
                \textbf{Discriminative}                                             \\
                \hline
                \citet{DBLP:conf/nips/VinyalsKKPSH15} & -    & -    & 88.3          \\
                \citet{DBLP:conf/acl/ZhuZCZZ13}       & 90.2 & 90.7 & 90.4          \\
                \citet{DBLP:conf/emnlp/CrossH16}      & 90.5 & 92.1 & 91.3          \\
                \citet{DBLP:journals/tacl/LiuZ17}     & 91.3 & 92.1 & 91.7          \\
                \citet{DBLP:conf/acl/SternAK17}       & 90.3 & 93.2 & 91.8          \\
                \citet{DBLP:journals/tacl/LiuZ17a}    & -    & -    & 91.8          \\
                \citet{DBLP:conf/acl/BengioSCJLS18}   & 92.0 & 91.7 & 91.8          \\
                \citet{DBLP:conf/acl/HongH18}         & 91.5 & 92.5 & 92.0          \\
                \citet{DBLP:conf/coling/TengZ18}      & 92.2 & 92.5 & 92.4          \\
                \citet{DBLP:conf/acl/KleinK18}        & 93.2 & 93.9 & \textbf{93.6} \\
                \hline
                Our in-order parser                   & 91.1 & 93.0 & 92.0          \\
                \hline\hline
                \textbf{Generative}                                                 \\
                \hline
                \citet{DBLP:conf/naacl/DyerKBS16}     & -    & -    & 89.8          \\
                \citet{DBLP:conf/emnlp/SternFK17}     & 92.5 & 92.5 & 92.5          \\
                \hline
            \end{tabular}
        \end{center}
        \caption{ Final results on the PTB test set. Here we only compare with single model parsers trained without external data.}
        \label{Tab:FinalResults}
    \end{table}

    \begin{table*}[t]
        \normalsize
        \begin{center}
            \begin{tabular}{l|l}
                \hline
                Sent (Line 5)    & ... \emph{step up to the plate to support the beleaguered floor traders} ...                                                                                                                                                                                  \\
                \hline
                Gold             & ... (VP \emph{step} (ADVP \emph{up} (PP \emph{to} (NP \emph{the} \emph{plate}))) (S (VP \emph{to} (VP \emph{support} ...)))) ...                                                                                                                              \\
                CKY-style        & ... (VP \emph{step} \textcolor[rgb]{1.00,0.00,0.00}{\emph{up}} (PP \emph{to} (NP \emph{the} \emph{plate})) (S (VP \emph{to} (VP \emph{support} ...)))) ...                                                                                                    \\
                Top-down         & ... (VP \emph{step} (ADVP \emph{up} (PP \emph{to} (NP \emph{the plate}))) (S (VP \emph{to} (VP \emph{support} ...)))) ...                                                                                                                                     \\
                In-order         & ... (VP \emph{step} (ADVP \emph{up} (PP \emph{to} (NP \emph{the plate}))) (S (VP \emph{to} (VP \emph{support} ...)))) ...                                                                                                                                     \\
                \hline
                Sent (Line 1090) & ... \emph{they 've really got to make the investment in people} ...                                                                                                                                                                                           \\
                \hline
                Gold             & ... (NP \emph{they}) (VP \emph{'ve} (ADVP \emph{really}) (VP \emph{got} (S (VP \emph{to} (VP \emph{make} (NP ...)))))) ...                                                                                                                                    \\
                CKY-style        & ... (NP \emph{they}) (VP \emph{'ve} (ADVP \emph{really}) (VP \emph{got} (S (VP \emph{to} (VP \emph{make} (NP ...)))))) ...                                                                                                                                    \\
                Top-down         & ... (NP \emph{they}) (VP \emph{'ve} \textcolor[rgb]{1.00,0.00,0.00}{(VP} (ADVP \emph{really}) \emph{got} (S (VP \emph{to} (VP \emph{make} \textcolor[rgb]{1.00,0.00,0.00}{(S} ...\textcolor[rgb]{1.00,0.00,0.00}{)})))\textcolor[rgb]{1.00,0.00,0.00}{)}) ... \\
                In-order         & ... (NP \emph{they}) (VP \emph{'ve} (ADVP \emph{really}) (VP \emph{got} (S (VP \emph{to} (VP \emph{make} (NP ...)))))) ...                                                                                                                                    \\
                \hline
                Sent (Line 1938) & \emph{They could still panic and bail out of the market} .                                                                                                                                                                                                    \\
                \hline
                Gold             & ... (VP (VP \emph{panic}) \emph{and} (VP \emph{bail} (PRT \emph{out}) (PP \emph{of} (NP \emph{the market})))) ...                                                                                                                                             \\
                CKY-style        & ... (VP \textcolor[rgb]{1.00,0.00,0.00}{\emph{panic}} \emph{and bail} \textcolor[rgb]{1.00,0.00,0.00}{(PP} \emph{out} (PP \emph{of} (NP \emph{the market}))\textcolor[rgb]{1.00,0.00,0.00}{)}) ...                                                            \\
                Top-down         & ... (VP (VP \emph{panic}) \emph{and bail} \textcolor[rgb]{1.00,0.00,0.00}{(PP} \emph{out} (PP \emph{of} (NP \emph{the market}))\textcolor[rgb]{1.00,0.00,0.00}{)}) ...                                                                                        \\
                In-order         & ... (VP (VP \emph{panic}) \emph{and bail} \textcolor[rgb]{1.00,0.00,0.00}{(PP} \emph{out} (PP \emph{of} (NP \emph{the market}))\textcolor[rgb]{1.00,0.00,0.00}{)}) ...                                                                                        \\
                \hline
            \end{tabular}
        \end{center}
        \caption{ Predictions of the three chart-based parsers on three examples from the test corpus. Red words represent the wrong predictions.}
        \label{Tab:examples}
    \end{table*}

    Next, we evaluate different configurations of the LSTM encoding history decisions as shown in Table \ref{Tab:CompDiffConfig}. First, we compare the results in line 2 and 6. We observe that span representations have better results than label embeddings as the input of the LSTM. The same results can be seen in line 3 and 7, which means that span representations are more important for the encoding on history decisions. Then for the output of the LSTM, we find that span predictions do not benefit much from history decisions as shown in line 2 and 3. The best result occurs in line 5 which uses both label embeddings and span representations as input and utilizes the output to predict labels and spans.

    The final test results are shown in Table \ref{Tab:FinalResults}. We achieve competitive results compared with other discriminative single models trained without external data. The best result from \citet{DBLP:conf/acl/KleinK18} is obtained by a self-attentive encoder and the second best one from \citet{DBLP:conf/coling/TengZ18} is obtained by local predictions on spans and CFG rules. Additionally, \citet{DBLP:conf/emnlp/SternFK17} get a better result than ours by generative methods.

    Furthermore, we use the pre-trained BERT \citep{DBLP:journals/corr/abs-1810-04805} to improve our in-order parser by simplely concatenating initial input embeddings and BERT vectors. We do not use fine-tuned BERT since DyNet has no ready-made BERT implementation, and we will implement it in future work. Despite this, the result has an improvement of 0.7 (92.7 F1) compared with the single model. The best result using fine-tuned BERT is \citet{DBLP:journals/corr/abs-1812-11760} which uses the same model as \citet{DBLP:conf/acl/KleinK18}. They use Transformer \citep{DBLP:conf/nips/VaswaniSPUJGKP17} rather than LSTM as the encoder and obtains a pretty high F1 score of 95.7, which shows the powerful encoding capability of Transformer.

    \subsection{Case Study}
    We select three examples from the test corpus and show the predictions of the three chart-based parsers in Table \ref{Tab:examples}.

    Given the 5-th sentence, only the CKY-style parser gives the wrong prediction. We observe that CKY-style parser combines the phrase ``\emph{set up}'' and the following phrases to VP because of its post-order traversal. However, top-down and in-order parser firstly assign ADVP to the word ``\emph{up}'' and then combine it with the following phrases. This shows that in-order parser can make the best of the lookahead information of top-down parser.

    Given the 1090-th sentence, top-down parser predicts incorrectly. It generates VP for the phrase ``\emph{really got to} ...'' firstly. However, this phrase should be combined with the word ``\emph{'ve}'' together, not individually be reduced to VP. CKY-style parser and in-order parser give the right predictions for that they do not give a non-terminal to the phrase ``\emph{really got to} ...'' at the beginning and reduce the whole phrase after all the phrases inside it being reduced. This shows that our in-order parser can better utilize the rich features from the local information than top-down parser.

    Given the 1938-th sentence, all of the three parsers give the wrong predictions. The problems of top-down parser and in-order parser are the same, which thinking of the words ``\emph{panic}'' and ``\emph{bail}'' as juxtaposed verbs and both modified by prepositional phrases ``\emph{out of} ...''. This can only be solved by considering the whole phrase after the two verbs. However, CKY-style model also suffers from this problem and incorrectly predicts the label of the word ``\emph{panic}''. We can speculate that top-down parser and in-order parser may have better predictions on unary chains than CKY-style parser.

    \section{Related Work}
    \label{Sec:RelatedWork}
    During recent years, many efforts have been made to balance the performance and efficiency of the parsing models. Based on transition systems \citep{DBLP:conf/emnlp/ChenM14}, in-order model \citep{DBLP:journals/tacl/LiuZ17a} is proposed to integrate the rich features of bottom-up models \citep{DBLP:conf/acl/ZhuZCZZ13, DBLP:conf/emnlp/CrossH16} and lookahead information of top-down models \citep{DBLP:journals/tacl/LiuZ17, DBLP:conf/eacl/SmithDBNKK17}. In order to solve the exposure bias problem, \citet{DBLP:conf/emnlp/Fernandez-Gonzalez18} propose dynamic oracles for top-down and in-order models which achieve the best performance amongst transition-based models. Furthermore, \citet{DBLP:conf/acl/HongH18} develop a new transition-based parser searching over exponentially large space like CKY-style models using beam search and cube pruning and reduce the time complexity to linear.

    Chart-based models usually have better performance but lower efficiency than transition-based models. Different from action sequence predictions in transition-based models, CKY-style and top-down inferences are applied to chart-based models based on independent scoring of labels and spans \citep{DBLP:conf/acl/SternAK17, DBLP:conf/naacl/GaddySK18}. However, CKY-style models have a high time complexity of $O(n^3)$ and top-down models cannot take all the states into account which will lose some performance. \citet{DBLP:journals/tacl/VieiraE17} improve the CKY-style models both on accuracy and runtime by learning pruning policies. Besides, self-attentive encoder is also used to obtain better lexical representations and achieves the best result so far \citep{DBLP:conf/acl/KleinK18}. Our model is inspired by \citet{DBLP:journals/tacl/LiuZ17a} and we also parse the constituent trees in the order of in-order traversal which can efficiently make the best of lookahead information and local information at the same time. Besides, we use LSTM to encode the history decisions and use the outputs to better predict the future labeled spans.

    In future work, we will replace the LSTM encoder with more powerful Transformer and try fine-tuned BERT to further improve the performance.

    \section{Conclusion}
    We propose a novel in-order chart-based constituent parsing model which utilizes the information of history decisions to improve the performance. The model not only achieves a high efficiency of the top-down models, but also performs better than the CKY-style models. Besides, we argue that history decisions are indeed helpful to the decoder of the top-down and in-order models. Our model achieves competitive results amongst the discriminative single models and is superior to previous chart-based models.

    %

    \newpage
    \bibliography{conll-2019}
    \bibliographystyle{acl_natbib}

\end{CJK*}
\end{document}